\documentclass[10pt,twocolumn,letterpaper]{article}

\usepackage{cvpr}
\usepackage{times}
\usepackage{epsfig}
\usepackage{graphicx}
\usepackage{amsmath}
\usepackage{amssymb}
\usepackage{breqn}
\usepackage{multirow}
 \usepackage{multicol}

 \usepackage[T1]{fontenc}
 \usepackage[utf8x]{inputenc}

\usepackage[]{algorithm}
\usepackage[noend]{algpseudocode}

\makeatletter
\def\BState{\State\hskip-\ALG@thistlm}
\makeatother

\newcommand{\Pexp}{\mathbin{\text{$\vcenter{\hbox{\textcircled{$+$}}}$}}}

\newcommand{\sym}{\mathbin{\text{$\vcenter{\hbox{\textcircled{$*$}}}$}}}

\usepackage[pagebackref=true,breaklinks=true,letterpaper=true,colorlinks,bookmarks=false]{hyperref}

\cvprfinalcopy 

\def\cvprPaperID{3451} 

\ifcvprfinal\fi
\begin{document}

\title{Differential Attention for Visual Question Answering}

\author{Badri Patro, Vinay P. Namboodiri\\
IIT Kanpur\\
{\tt\small \{ badri,vinaypn \}@iitk.ac.in}
}

\maketitle

\begin{abstract}
In this paper we aim to answer questions based on images when provided with a dataset of question-answer pairs for a number of images during training. A number of methods have focused on solving this problem by using image based attention. This is done by focusing on a specific part of the image while answering the question. Humans also do so when solving this problem. However, the regions that the previous systems focus on are not correlated with the regions that humans focus on. The accuracy is limited due to this drawback. In this paper, we propose to solve this problem by using an exemplar based method. We obtain one or more supporting and opposing exemplars to obtain a differential attention region. This differential attention is closer to human attention than other image based attention methods. It also helps in obtaining improved accuracy when answering questions. The method is evaluated on challenging benchmark datasets. We perform better than other image based attention methods and are competitive with other state of the art methods that focus on both image and questions.

\end{abstract}

\section{Introduction}
Answering questions regarding images requires us to obtain an understanding about the image. We can gain insights into a method by observing the region of an image the method focuses on while answering a question. It has been observed in a recent work that humans also attend to specific regions of an image while answering questions \cite{Das_EMNLP2016}. We therefore expect strong correlation between focusing on the ``right’’ regions while answering questions and obtaining better semantic understanding to solve the problem. This correlation exists as far as humans are concerned~\cite{Das_EMNLP2016}. We therefore aim in this paper to obtain image based attention regions that correlate better with human attention. We do that by obtaining a differential attention. The differential attention relies on an exemplar model of cognition.
\begin{figure}[ht]
	\centering
	\includegraphics[width=0.45\textwidth]{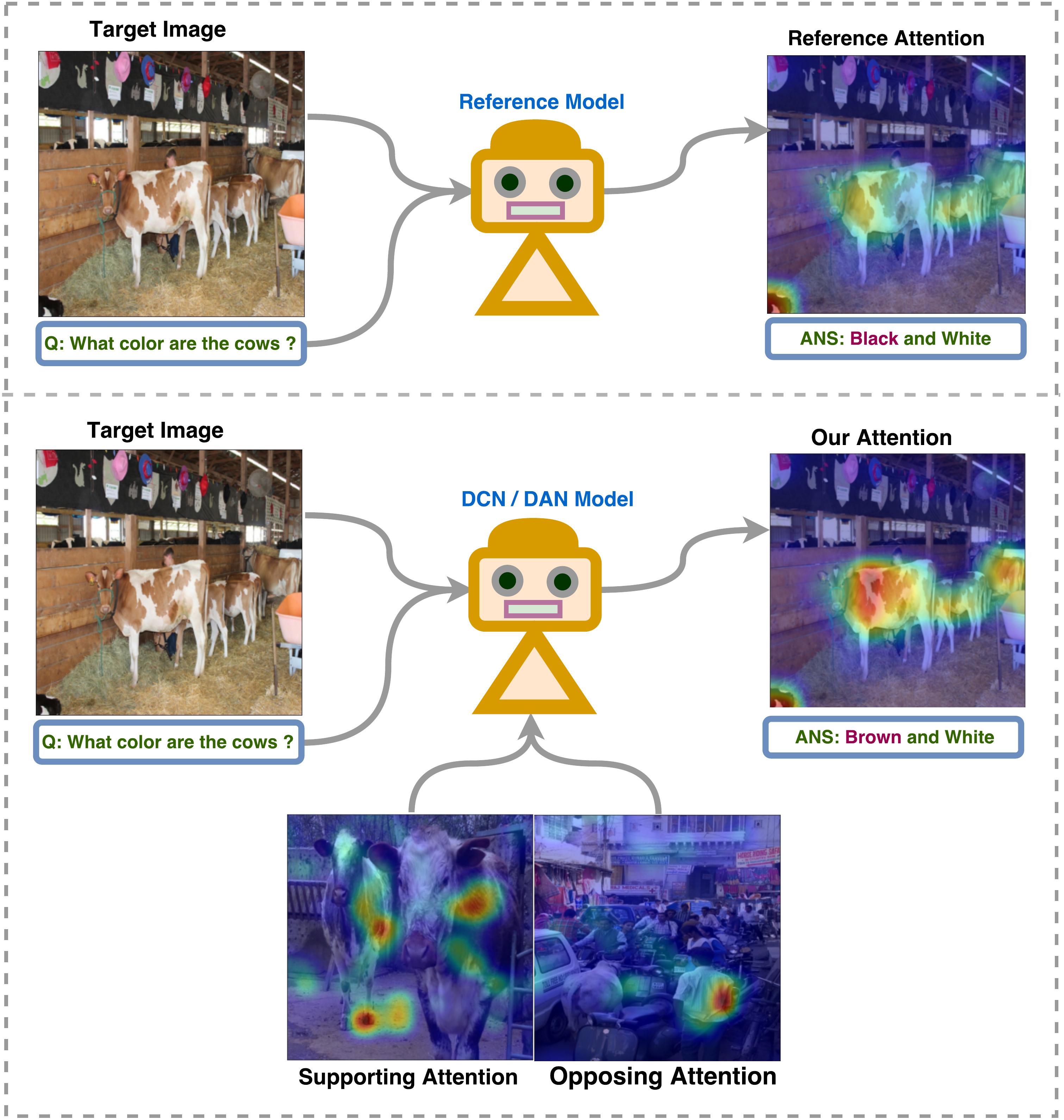}
	\caption{ Illustration of improved attention obtained using Differential Context Network. Using the baseline reference we got answer as: ``Black and White'' But, using our methods DAN or DCN we get answer as ``Brown and White'', that is actually the color of the cow. We provide the attention map that indicates the actual improvement in attention.}
	\label{fig:flow}
\end{figure}

In cognition studies, the exemplar theory suggests that humans are able to obtain generalisation for solving cognitive tasks by relying on an exemplar model. In this model, individuals compare new stimuli with the instances already stored in memory \cite{Jakel_2008}\cite{Shepard_Science1987} and obtain answers based on these exemplars. We would like an exemplar model to provide attention. We want to focus on the specific parts in a nearest exemplar that distinguishes it from a far example. We do that by obtaining the differential attention region that distinguishes a supporting exemplar from an opposing exemplar. Our premise is that the difference between a nearest semantic exemplar and a far semantic exemplar can guide attention on a specific image region. We show that by using this differential attention mechanism we are able to obtain significant improvement in solving the visual question answering task. Further, we show that the obtained attention regions are more correlated with human attention regions both quantitatively and qualitatively. We evaluate this on the challenging VQA-1 \cite{VQA}, VQA-2 \cite{Goyal_CVPR2017} and HAT \cite{Das_EMNLP2016}

The main flow of the method followed is outlined in figure~\ref{fig:flow}. Given an image and an associated question, we use an attention network to combine the image and question to obtain a reference attention embedding. This is used to order the examples in the database. We obtain a near example as a supporting exemplar and a far example as an opposing exemplar. These are used to obtain a differential attention vector. We evaluate two variants of our approach, one we term as `differential attention network' (DAN) where the supporting and opposing exemplars are used only to provide a better attention on the image. The other we term as `differential context network' (DCN) that obtains a differential context feature. This is obtained from the difference between supporting  and opposing exemplars with respect to the original image to provide a differential feature. The additional context is used in answering questions. Both variants improve results over the baseline with the differential context variant being better.

Through this paper we provide the following contributions
\begin{itemize}
    \item We adopt an exemplar based approach to improve visual question answering (VQA) methods by providing a differential attention
    \item We evaluate two variants for obtaining differential attention - one where we only obtain attention and the other where we obtain differential context in addition to attention
    \item We show that this method correlates better with human attention and results in an improved visual question answering that improves the state-of-the-art for image based attention methods. It is also competitive with respect to other proposed methods for this problem.
\end{itemize}

\section{Related Work}
\label{sec:lit_surv}

The problem of Visual Question Answering (VQA) is a recent problem that was initiated as a new kind of visual Turing test. The aim was to show progress of systems in solving even more challenging tasks as compared to the traditional visual recognition tasks such as object detection and segmentation. An initial work in this area was by Geman {\it et al.} \cite{Geman_PNAS2015} that proposed this visual Turing test. Around the same time Malinowski {\it et al.} \cite{Malinowski_NIPS2014} proposed a multi-world based approach to obtain questions and answer them from images. These works aimed at answering questions of a limited type. In this work we aim at answering free-form open-domain\cite{VQA} questions as was attempted by later works.

An initial approach towards solving this problem in the open-domain form was by  \cite{Malinowski_ICCV2015}. This was inspired by the work on neural machine translation that proposed translation as a sequence to sequence encoder-decoder framework \cite{Sutskever_NIPS2014}. However, subsequent works \cite{Ren_NIPS2015}\cite{VQA} approached the problem as a classification problem using encoded embeddings. They used soft-max classification over an image embedding  (obtained by a CNN) and a question embedding (obtained using an LSTM). Further work by Ma {\it et al.} \cite{Ma_AAAI2016} varied the way to obtain an embedding by using CNNs to obtain both image and question embeddings. Another interesting approach \cite{Noh_CVPR2016} used dynamic parameter prediction where weights of the CNN model for the image embedding are modified based on the question embedding using hashing.  These methods however, are not attention based. Use of attention enables us to focus on specific parts of an image or question that are pertinent for an instance and also offer valuable insight into the performance of the system.

There has been significant interest in including attention to solve the VQA problem. Attention based models comprises of image based attention models, question based attention and some that are both image and question based attention. In image based attention approach the aim is to use the question in order to focus attention over specific regions in an image \cite{Shih_CVPR2016}. An interesting recent work \cite{Yang_CVPR2016} has shown that it is possible to repeatedly obtain attention by using stacked attention over an image based on the question. Our work is closely related to this work. There has been further work \cite{Li_NIPS2016} that considers a region based attention model over images. The image based attention has allowed systematic comparison of various methods as well as enabled analysis of the correlation with human attention models as shown by \cite{Das_EMNLP2016}.  In our approach we focus on image based attention using differential attention and show that it correlates better with image based attention. There has been a number of interesting works on question based attention as well \cite{Zhu_CVPR2016}\cite{Xu_ECCV2016}. An interesting work obtains varied set of modules for answering questions of different types \cite{andreas16naacl}. Recent work also explores joint image and question based hierarchical co-attention \cite{Lu_NIPS2016}. The idea of differential attention can also be explored through these approaches.  However, we restrict ourselves to image based attention as our aim is to obtain a method that correlates well with human attention~\cite{Das_EMNLP2016}. There has been an interesting work by \cite{Fukui_arXiv2016} that advocates multimodal pooling and obtains state of the art in VQA. Interestingly, we show that by combining it with the proposed method further improves our results.

\section{Method}
\label{sec:method}
\begin{figure*}[ht]
	\centering
	\includegraphics[width=0.9\textwidth,height=0.35\textwidth]{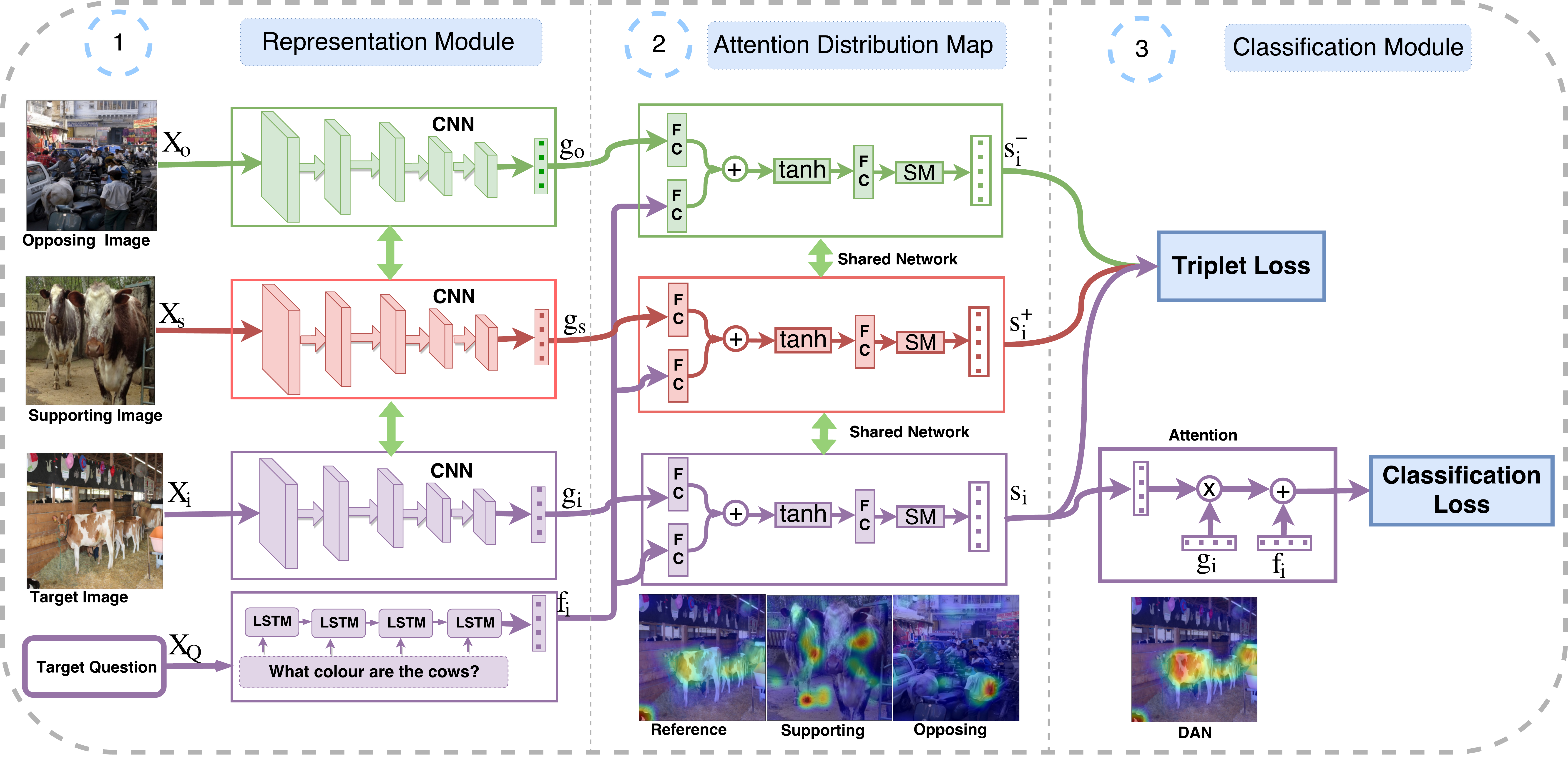}
	\caption{ Differential Attention Network}
	\label{fig:DAN}
\end{figure*}

In this paper we adopt a classification framework that uses the image embedding combined with the question embedding to solve for the answer using a softmax function in a multiple choice setting. A similar setting is adopted in the Stacked Attention Network (SAN) \cite{Yang_CVPR2016}, that also aims at obtaining better attention and several other state-of-the-art methods. We provide two different variants for obtaining differential attention in the VQA system. We term the first variant a `Differential Attention Network' (DAN) and the other a `Differential Context Network' (DCN). We explain both the methods in the following sub-sections. A common requirement for both these tasks is to obtain nearest semantic exemplars. 

\subsection{Finding Exemplars}
In our method, we use semantic nearest neighbors. Image level similarity does not suffice as the nearest neighbor may be visually similar but may not have the same context implied in the question (for instance, `Are the children playing?' produces similar results for images with children based on visual similarity, whether the children are playing or not). In order to obtain semantic features we use a VQA system \cite{Lu2015} to provide us with a joint image-question level embedding that relates meaningful exemplars. We compared image level features against the semantic nearest neighbors and observed that the semantic nearest neighbors were better. We used the semantic nearest neighbors in a k-nearest neighbor approach using a K-D tree data structure to represent the features. The ordering of the data-set features is based on the Euclidean distance. In section~\ref{sec:param_analysis} we provide the evaluation with several values of nearest neighbors that were used as supporting exemplars. For obtaining opposing exemplar we used a far neighbor that was an order of magnitude further than the nearest neighbor. This we obtained through a coarse quantization of training data into bins. We specified the opposing exemplar as one that was around 20 clusters away in a 50 cluster ordering. This parameter is not stringent and it only matters that the opposing exemplar be far from the supporting exemplar. We show that using these supporting and opposing exemplars aids the method and any random ordering adversely effects the method.

\subsection{Differential Attention Network (DAN)}
In the DAN method, we use a multi-task setting. As one of the tasks we use a triplet loss\cite{Hoffer_Springer2015} to learn a distance metric. This metric ensures that the distance between the attention weighted regions of near examples is less and the distance between attention weighted far examples is more. The other task is the main task of VQA.
More formally, given an image $x_i$ we obtain an embedding $g_i$ using a CNN that we parameterize through a function $G(x_i,W_c)$ where $W_c$ are the weights of the CNN. Similarly the question $q_i$ results in a question embedding $f_i$ after passing through an LSTM parameterised using the function $F(q_i, W_l)$ where $W_l$ are the weights of the LSTM. This is illustrated in part 1 of figure~\ref{fig:DAN}. The output image embedding $g_i$ and question embedding $f_i$ are used in an attention network that combines the image and question embeddings with a weighted softmax function and produces an output attention weighted vector $s_i$. The attention mechanism is illustrated in figure~\ref{fig:DAN}. The weights of this network are learnt end-to-end learning using the two losses, a triplet loss and a soft-max classification loss for the answer (shown in part 3 of figure~\ref{fig:DAN}).  The aim is to obtain attention weight vectors that bring the supporting exemplar attention close to the image attention and far from the opposing exemplar attention. The joint loss function used for training is given by:


\begin{figure*}[ht]
	\centering
	\includegraphics[width=0.9\textwidth]{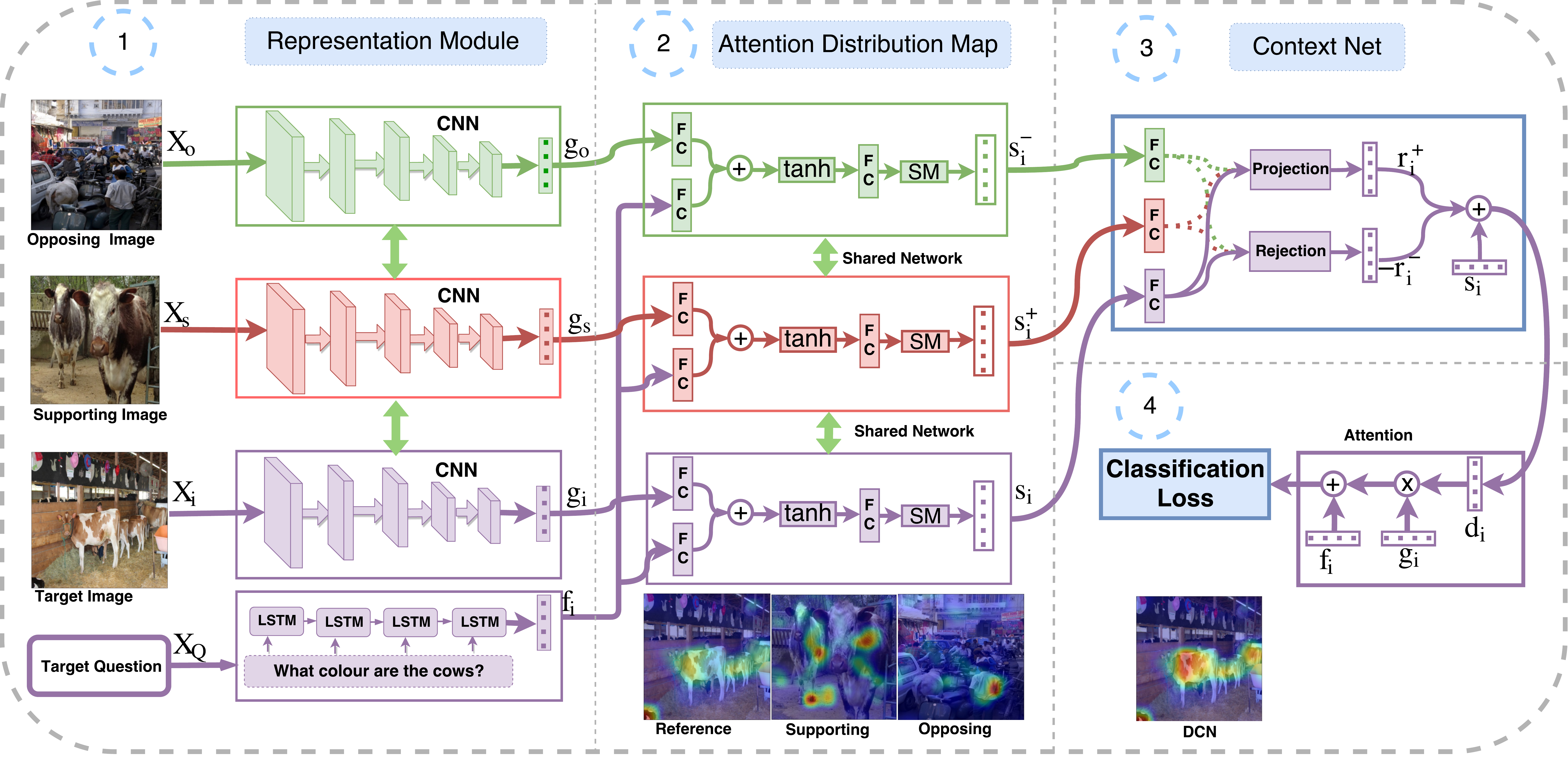}
	\caption{ Differential Context Network}
	\label{fig:DCN}
\end{figure*}

\begin{equation}
\begin{split}
&    L( \textbf{s},\textbf{y},\theta) = \frac{1}{N}\sum^{N}_{i=1}\left( L_{cross}( \textbf{s},\textbf{y}) + \nu T(s_i,s_i^+, s_i^-) \right)\\
& L_{cross}( \textbf{s},\textbf{y}) = -\frac{1}{C}\sum_{j=1}^{C} y_{j} \texttt{log} \texttt{p}(c_{j}|\textbf{s}) 
    \end{split}
\end{equation}
where $\theta$ is the set of model parameters for the two loss functions, $y$ is the output class label and $s$ is the input sample. $C$ is the total number of classes in VQA ( consists of the set of total number of output classes including color, count etc. ) and  $N$ is the total number of samples. The first term is the classification loss and the second term is the triplet loss. $\nu$ is a constant that controls the ratio between classification loss and triplet loss.  $T(s_i, s_i^{+}, s_i^{-})$ is the triplet loss function that is used. This is decomposed into two terms, one that brings the positive sample closer and one that pushes the negative sample farther. This is given by
\begin{dmath}
T(s_i,s_i^+,s_i^-) = \texttt{max}(0, ||t(s_{i})-t(s_{i}^{+})||^{2}_{2} + \alpha - ||t(s_{i})-t(s_{i}^{-})||^{2}_{2})
\end{dmath}
The constant $\alpha$ controls the separation margin between supporting and opposing exemplars. The constants $\nu$ and $\alpha$ are obtained through validation data.

The method is illustrated in figure~\ref{fig:DAN}. We further extend the model to a quintuplet setting where we bring two supporting attention weights closer and two opposing attention weights further in a metric learning setting. We observe in section~\ref{sec:param_analysis} that this further improves the performance of the DAN method.

\begin{table*}[htb]
		\vspace{-0.3cm}
	\caption{Analysis network parameter for DAN }
	\label{tab:dan_param}
	\begin{center}
		
		\begin{tabular}{l|cccr | c } \hline
			
	\textbf{Models}&  \multicolumn{4}{c|}{VQA1.0 Open-Ended (test-dev)}&  HAT val dataset \\ 
	 \cline{2-6}
	 & \textbf{All} & \textbf{Yes/No} & \textbf{Number} & \textbf{others} & \textbf{Rank-correlation} \\ \hline 	
	 
			LSTM Q+I+ Attention(LQIA) & { 56.1 }& { 80.3 }& { 37.4 }& {  40.46}& { 0.2142} \\ \hline
			
			DAN(K=1)+LQIA  & {59.2 }& { 80.1 }& { 36.1 }& { 46.6}& { 0.2959} \\ 				
			DAN(K=2)+LQIA  & {59.5 }& { 80.9 }& { 36.6 }& { 47.1 }& { 0.3090} \\ 				
			DAN(K=3)+LQIA  & {59.9 }& { 80.6 }& { 37.2 }& \textbf{ 47.5 }& { 0.3100}\\ 
			DAN(K=4)+LQIA  & \textbf{60.2 }& \textbf{ 80.9 }& \textbf{ 37.4 }& { 47.2 } & \textbf{ 0.3206 }\\ \hline

            DAN(K=1)+MCB  & {64.8 }& { 82.4 }& { 38.1 }& { 54.2}& { 0.3284} \\ 				
            DAN(K=2)+MCB  & {64.8 }& { 82.9 }& { 38.0 }& { 54.3 }& { 0.3298} \\ 				
            DAN(K=3)+MCB  & {64.9 }& { 82.6 }& { 38.2 }& { 54.6 }& { 0.3316}\\ 
            DAN(K=4)+MCB  & \textbf{65.0}& { 83.1 }& \textbf{ 38.4 }& \textbf{ 54.9}& \textbf{ 0.3326} \\ \hline 

			DAN(K=5)+LQIA & {58.1 }& { 79.4 }& { 36.9 }& { 45.7 }& { 0.2157} \\ 
			DAN(K=1,Random)+LQIA & { 56.4 }& { 79.3 }& { 37.1 }& {  44.6} & {0.2545}\\ \hline
			
		\end{tabular}
		
	\end{center}
		\vspace{-0.3cm}
\end{table*}

\begin{table*}[htb]
	 	\vspace{-0.3cm}
	 	\caption{Analysis network parameter for DCN}
	 	\label{tab:dcn_param}
	 	\begin{center}
		
		\begin{tabular}{l|cccr | c }  \hline

		 \multirow{2}{*}{\textbf{Models}}&  \multicolumn{4}{c|}{VQA1.0 Open-Ended (test-dev)}&  HAT val dataset \\ 
		 \cline{2-6}
								& \textbf{All} & \textbf{Yes/No} & \textbf{Number} & \textbf{others} & \textbf{Rank-correlation} \\ \hline

			LSTM Q+I+ Attention(LQIA) & { 56.1 }& { 80.3 }& { 37.4 }& {  40.46} & {  0.2142}\\ \hline
			
			DCN Add\_v1(K=4)+(LQIA)      & {60.4 }& {81.0 }& { 37.5 }& { 47.1}& {0.3202} \\ 		
			DCN Add\_v2(K=4)+(LQIA) & {60.4 }& {81.2 }& { 37.2 }& { 47.3 } & {0.3215}\\ 
			DCN Mul\_v1(K=4)+(LQIA)      & {60.6 }& {80.9 }& \textbf{ 37.8 }& { 47.9 }& {0.3229} \\ 
			DCN Mul\_v2(K=4)+(LQIA)  & \textbf{60.9 }& \textbf{81.3 }& { 37.5 }& \textbf{ 48.2 }& \textbf{0.3242} \\ \hline
			
            DCN Add\_v1(K=4)+MCB  & {65.1}& { 83.1 }& { 38.5 }& { 54.5 }& {0.3359}\\
            DCN Add\_v2(K=4)+MCB  & {65.2}& { 83.4 }& { 39.0 }& { 54.6 }& {0.3376}\\
            DCN Mul\_v1(K=4)+MCB  & {65.2}& \textbf{ 83.9 }& { 38.7 }& { 54.9 }& {0.3365}\\
            DCN Mul\_v2(K=4)+MCB  & \textbf{65.4}& { 83.8 }& \textbf{ 39.1 }& \textbf{ 55.2 }& \textbf{0.3389}\\ \hline
		\end{tabular}
		
	\end{center}
		\vspace{-0.3cm}
\end{table*}
\subsection{Differential Context Network (DCN)}
We next consider the other variant that we propose where the differential context feature is added instead of only using it for obtaining attention. The first two parts are same as that for the DAN network. In part 1, we use the image, the supporting and the opposing exemplar and obtain the corresponding image and question embedding. This is followed by obtaining attention vectors $s_i, s_i^+, s_i^-$ for the image, the supporting and the opposing exemplar. While in DAN, these were trained using a triplet loss function, in DCN, we obtain two context features, the supporting context $r_i^+$ and the opposing context $r_i^-$. This is shown in part 3 in figure~\ref{fig:DCN}. The supporting context is obtained using the following equation

\begin{equation}
r_i^+ = (s_i \bullet s_i^+)\frac{s_i} {\lVert s_i \rVert_{L_{2}}^{2}} + (s_i \bullet s_i^-)\frac{s_i} {\lVert s_i \rVert_{L_{2}}^{2}}
\end{equation}


where $\bullet$ is the dot product. This results in obtaining correlations between the attention vectors. 

The first term of the supporting context $r_i^+$ is the vector projection of $s_i^+$ on  $s_i$ and and second term is the vector projection of $s_i^-$ on  $s_i$. Similarly, for opposing context we compute  vector projection of $s_i^+$ on  $s_i$ and $s_i^-$ on  $s_i$.  The idea is that the projection measures similarity between the vectors that are related. We subtract the vectors that are not related from the resultant. While doing so, we ensure that we enhance similarity and only remove the feature vector that is not similar to the original semantic embedding. This equation provides the additional feature that is supporting and is relevant for answering the current question $q_i$ for the image $x_i$.

Similarly, the opposing context is obtained by the following equation

\begin{equation}
r_i^- = (s_i^+ - (s_i \bullet s_i^+)\frac{s_i} {\lVert s_i \rVert_{{2}}^{2}}) + (s_i^- - (s_i \bullet s_i^-)\frac{s_i} {\lVert s_i \rVert_{{2}}^{2}})
\end{equation}

We next compute the difference between the supporting and opposing context features i.e. $r_i^+ - r_i^-$ that provides us with the differential context feature $\hat{d_i}$. This is then either added with the original attention vector (DCN-Add) or multiplied with the original attention vector $s_i$ (DCN-Mul) providing us with the final differential context attention vector $d_i$. This is then the final attention weight vector multiplied to the image embedding $g_i$ to obtain the vector $v_i$ that is then used with the classification loss function. This is shown in part 4 in the figure~\ref{fig:DCN}. The resultant attention is observed to be better than the earlier differential attention feature obtained through DAN as the features are also used as context.

The network is trained end-to-end using the following soft-max classification loss function
\begin{equation}
    L( \textbf{v},\textbf{y},\theta) = -\sum_{j=1}^{C} y_{j} \texttt{log} \texttt{p}(c_{j}|\textbf{v}) 
\end{equation}



\section{Experiments}
\label{sec:results}

	\begin{table*}[ht]
		\caption{Open-Ended VQA1.0 accuracy on test-dev}
		\label{VQA1_accuracy}
		\begin{center}			
			\begin{tabular}{lcccr  } \hline
				
				\textbf{Models} & \textbf{All} & \textbf{Yes/No} & \textbf{Number} & \textbf{others}  \\ \hline 
				LSTM Q+I \cite{VQA}& { 53.7 }& {78.9 }& { 35.2}& { 36.4} \\
			    LSTM Q+I+ Attention(LQIA) & { 56.1 }& { 80.3 }& { 37.4 }& {  40.4} \\
				DPPnet \cite{Noh_CVPR2016}    & { 57.2 }& {80.7 }& { 37.2 }& { 41.7} \\
			 	SMem \cite{Xu_ECCV2016}       & { 58.0 }& {80.9 }& { 37.3 }& { 43.1} \\ 
				SAN \cite{Yang_CVPR2016}      & { 58.7 }& {79.3 }& { 36.6 }& { 46.1 }\\
				QRU(1)\cite{Li_NIPS2016}  & { 59.3 }& {81.0 }& { 35.9 }& { 46.0 }\\\hline
			
				DAN(K=4)+ LQIA & \textbf{60.2 }& { 80.9 }& \textbf{ 37.4 }& \textbf{ 47.2 } \\ \hline
				DMN+\cite{Xiong_arXiv2016} & {60.3 }& { 80.5 }& { 36.8 }& { 48.3 } \\ 
				QRU(2)\cite{Li_NIPS2016}  & { 60.7 }& {82.3 }& { 37.0 }& { 47.7 }\\\hline				
				DCN Mul\_v2(K=4)+LQIA  & \textbf{60.9 }& {81.3 }& \textbf{ 37.5 }& \textbf{ 48.2 } \\ \hline

				HieCoAtt \cite{Lu_NIPS2016} & {61.8}& { 79.7 }& { 38.9 }& { 51.7 } \\
				
				MCB + att \cite{Fukui_arXiv2016} & {64.2}& { 82.2 }& { 37.7 }& { 54.8 } \\
					
				MLB \cite{Kim_ICLR2017} & {65.0}& \textbf{ 84.0 }& { 37.9 }& { 54.7 } \\ \hline
                DAN(K=4)+ MCB  & \textbf{65.0}& { 83.1 }& \textbf{ 38.4 }& \textbf{ 54.9} \\ 
                DCN Mul\_v2(K=4)+MCB  & \textbf{65.4}& { 83.8 }& \textbf{ 39.1 }& \textbf{ 55.2 } \\ \hline
				
			\end{tabular}
		\end{center}		
	\end{table*}
	
	\begin{table*}[ht]
			\caption{VQA2.0 accuracy on Validation set for DCN and DAN }
			\label{VQA2_accuracy}
			\begin{center}
				
				\begin{tabular}{lcccr  } \hline
					
					\textbf{Models} & \textbf{All} & \textbf{Yes/No} & \textbf{Number} & \textbf{others}  \\ \hline 
					SAN-2           & {52.82 }& {- }& { - }& { - }\\  \hline 
					DAN(K=1) +LQIA        &\textbf {52.96}& \textbf{70.08 }& \textbf{34.06 }& \textbf{ 44.20} \\  \hline
					
					DCN Add\_v1(K=1)+LQIA     & {53.01 }& { 70.13 }& { 33.98 }& { 44.27} \\ 
					DCN Add\_v2(K=1) +LQIA& {53.07 }& { 70.46 }& { 34.30 }& { 44.10 } \\ 					
					DCN Mul\_v1(K=1) +LQIA    & {53.18 }& { 70.24 }& { 34.53 }& { 44.24 } \\ 					
					DCN Mul\_v2(K=1)+LQIA & \textbf{53.26}& \textbf{ 70.57 }& \textbf{ 34.61}& \textbf{ 44.39 } \\ \hline

					DCN Add\_v1(K=4)+MCB & {65.30 }& { 81.89 }& { 42.93 }& { 55.56 } \\ 
					DCN Add\_v2(K=4)+MCB & {65.41 }& { 81.90 }& { 42.88 }& { 55.99 } \\ 					
					DCN Mul\_v1(K=4)+MCB & {65.52 }& { 82.07 }& { 42.91 }& { 55.97 } \\ 					
					DCN Mul\_v2(K=4)+MCB & \textbf{65.90}& \textbf{ 82.40 }& \textbf{ 43.18}& \textbf{ 56.81 } \\ \hline

				\end{tabular}
				
			\end{center}
			
		\end{table*}

The experiments have been conducted using the two variants of differential attention that are proposed and compared against baselines on standard datasets. We first analyze the different parameters for the two variants DAN and DCN that are proposed. We further evaluate the two networks by comparing the networks with comparable baselines and evaluate the performance against the state of the art methods. The main evaluation is conducted to evaluate the performance in terms of correlation of attention with human correlation where we obtain state-of-the-art in terms of correlation with human attention. Further, we observe that its performance in terms of accuracy for solving the VQA task is substantially improved and is competitive with the current state of the art results on standard benchmark datasets. We also analyse the performance of the network on the recently proposed VQA2 dataset. 

\subsection{Analysis of Network Parameters}
\label{sec:param_analysis}
In the proposed DAN network, we have a dependency on the number of k-nearest neighbors that should be considered. We observe in table~\ref{tab:dan_param}, that using 4 nearest neighbors in the triplet network we obtain the highest correlation with human attention as well as accuracy using VQA-1 dataset. We therefore use 4 nearest neighbors in our experiments. We observe that increasing nearest neighbors beyond 4 nearest neighbors results in reduction in accuracy. Further, even using a single nearest neighbor results in substantial improvement that is marginally improved as we move to 4 nearest neighbors. 

We also evaluate the effect of using the nearest neighbors as obtained through a baseline model~\cite{{VQA}} versus using a random assignment of supporting and opposing exemplar. We observe that using DAN with a random set of nearest neighbors decreases the performance of the network. While comparing the network parameters, the comparable baseline we use is the basic model for VQA using LSTM and CNN \cite{{VQA}}. This however does not use attention and we evaluate this method with attention. With the best set of parameters the performance improves the correlation with human attention by 10.64\%. We also observe that correspondingly the VQA performance improves by 4.1\% over the comparable baseline. We further then incorporate this model with the model from MCB \cite{Fukui_arXiv2016} which is a state of the art VQA model. This further improves the result by 4.8\% more on VQA and a further increase in correlation with human attention by 1.2\%. 
 	
 	
 In the proposed DCN network we have two different configurations, one where we use the add module (DCN-add) for adding the differential context feature and one where we use the (DCN-mul) multiplication module for adding the differential context feature. We further have a dependency on the number of k-nearest neighbors for the DCN network as well. This is also considered. We next evaluate the effect of using a fixed scaling weight (DCN\_v1) for adding the  differential context feature against learning a linear scaling weight (DCN\_v2) for adding the differential context feature. All these parameter results are compared in table~\ref{tab:dcn_param}.

 As can be observed from table~\ref{tab:dcn_param} the configuration that obtains maximum accuracy on VQA dataset \cite{VQA} and in correlation with human attention is the version that uses multiplication with learned weights and with 4 nearest neighbors being considered. This results in an improvement of 11\% in terms of correlation with human attention and 4.8\% improvement in accuracy on the VQA-1 dataset \cite{VQA}. We also observe that incorporating DCN with MCB \cite{Fukui_arXiv2016} further improves the results by 4.5\% further on VQA dataset and results in an improvement of 1.47\% improvement in correlation with attention. These configurations are used in comparison with the baselines. 
 \begin{table}[ht]
	\caption{Rank Correlation on HAT Validation Dataset for DAN and DCN}
	\label{HAT_rank_correlation}
	\begin{center}		
		\begin{tabular}{ lc } \hline
			
			\textbf{Models} & \textbf{Rank-correlation} \\ \hline 
			LSTM Q+I+ Attention(LQIA)	 			& {0.214 $\pm$ 0.001} \\
			SAN\cite{Das_EMNLP2016} 		& {0.249 $\pm$  0.004} \\			
			HieCoAtt-W\cite{Lu_NIPS2016}	& {0.246 $\pm$ 0.004} \\		
			HieCoAtt-P \cite{Lu_NIPS2016}	& {0.256 $\pm$ 0.004}\\
			HieCoAtt-Q\cite {Lu_NIPS2016}	& {0.264 $\pm$ 0.004}\\
			MCB + Att.	& {0.279 $\pm$ 0.004} \\\hline
			DAN (K=4) +LQIA	 					& { 0.321$\pm$ 0.001} \\
			DCN Mul\_v2(K=4) +LQIA				& \textbf{ 0.324$\pm$ 0.001}  \\ \hline
            DAN (K=4) +MCB & { 0.332$\pm$ 0.001} \\
            DCN Mul\_v2(K=4) +MCB & \textbf{ 0.338$\pm$ 0.001}  \\ \hline
			
			Human \cite{Das_EMNLP2016}   	& { 0.623 $\pm$ 0.003 } \\ \hline
		\end{tabular}	
	\end{center}
\end{table}

 \subsection{Comparison with baseline and state of the art}
We obtain the initial comparison with the baselines on the rank correlation on human attention (HAT) dataset \cite{Das_EMNLP2016} that provides human attention by using a region deblurring task while solving for VQA. Between humans the rank correlation is 62.3\%. The comparison of various state-of-the-art methods and baselines are provided in table~\ref{HAT_rank_correlation}. The baseline we use \cite{{VQA}} is the method used by us for obtaining exemplars. This uses a question embedding using an LSTM and an image embedding using a CNN. We additionally consider a variant of the same that uses attention. We have also obtained results for the stacked attention network~\cite{Yang_CVPR2016}. The results for the Hierarchical Co-Attention work~\cite{Lu_NIPS2016} are obtained from the results reported in Das {\it et al.}~\cite{Das_EMNLP2016}. We observe that in terms of rank correlation with human attention we obtain an improvement of around 10.7\% using DAN network (with 4 nearest neighbors) and using DCN network (4 neighbors with multiplication module and learned scaling weights) we obtain an improvement of around 11\% over the comparable baseline. We also obtain an improvement of around 6\% over the Hierarchical Co-Attention work \cite{Lu_NIPS2016} that uses co-attention on both image and questions. Further incorporating MCB improves the results for both DAN and DCN resulting  in an improvement of 7.4\% over Hierarchical co-attention work and 5.9\% improvement over MCB method. However, as noted by \cite{Das_EMNLP2016}, using a saliency based method  \cite{Judd_ICCV2009} that is trained on eye tracking data to obtain a measure of where people look in a task independent manner results in more correlation with human attention (0.49). However, this is explicitly trained using human attention and is not task dependent. In our approach, we aim to obtain a method that can simulate human cognitive abilities for solving tasks.

	
	

	\begin{figure}[ht]
	\centering
	\includegraphics[width=0.4\textwidth]{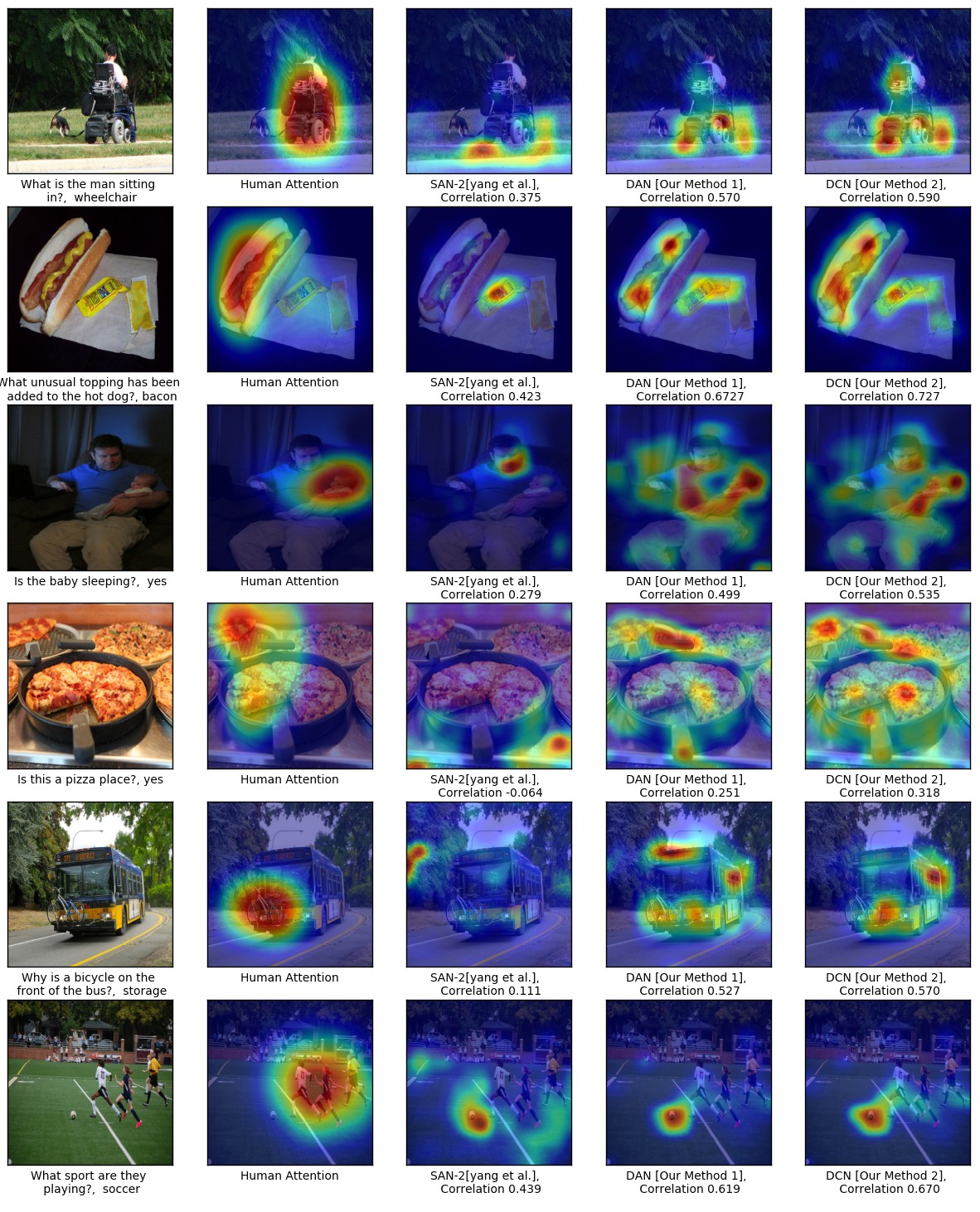}
	\caption{ In this figure, the first column indicates target question and corresponding image, second column indicates reference human attention map in HAT dataset, third column refer to generated attention map for SAN, fourth column refers to rank correlation of our DAN model and final column refers to rank correlation for our DCN model.}
	\label{fig:att}
		\vspace{-0.8cm}
\end{figure}

	
	
We next evaluate the different baseline and state of the art methods on the VQA dataset \cite{VQA} in table~\ref{VQA1_accuracy}. There have been a number of methods proposed for this benchmark dataset for evaluating the VQA task. Among the notable different methods, the Hierarchical Co-Attention work \cite{Lu_NIPS2016} obtains 61.8\% accuracy on VQA task, the  dynamic parameter prediction \cite{Noh_CVPR2016} method obtains 57.2\% and the stacked attention network \cite{Yang_CVPR2016} obtains 58.7\% accuracy. We observe that the differential context network performs well outperforming all the image based attention methods and results in an accuracy of 60.9\%. This is a strong result and we observe that the performance improves across different kinds of questions. Further, on combining the method with MCB, we obtain improved results of 65\% and 65.4\% using  DAN and DCN respectively improving over the results of MCB  by 1.2\%. This is consistent with the improved correlation with human attention that we observe in table~\ref{HAT_rank_correlation}.
%
%
%

We next evaluate the proposed method on a recently proposed VQA-2 dataset \cite{Goyal_CVPR2017}. The aim in this new dataset is to remove the bias in different questions. It is a more challenging dataset as compared to the previous VQA-1 dataset \cite{VQA}. We provide a comparison of the proposed DAN and DCN methods against the stacked attention network (SAN) \cite{Yang_CVPR2016} method. As can be observed in table~\ref{VQA2_accuracy}, the proposed methods obtain improved performance over a strong stacked attention baseline. We observe that our proposed methods are also able to improve the result over the SAN method. DCN with 4 nearest neighbors when combined with MCB obtains an accuracy of 65.90\%

. 
		


	\begin{figure*}[ht]
\vspace{-0.3cm}
	\centering
	\includegraphics[width=0.85\textwidth]{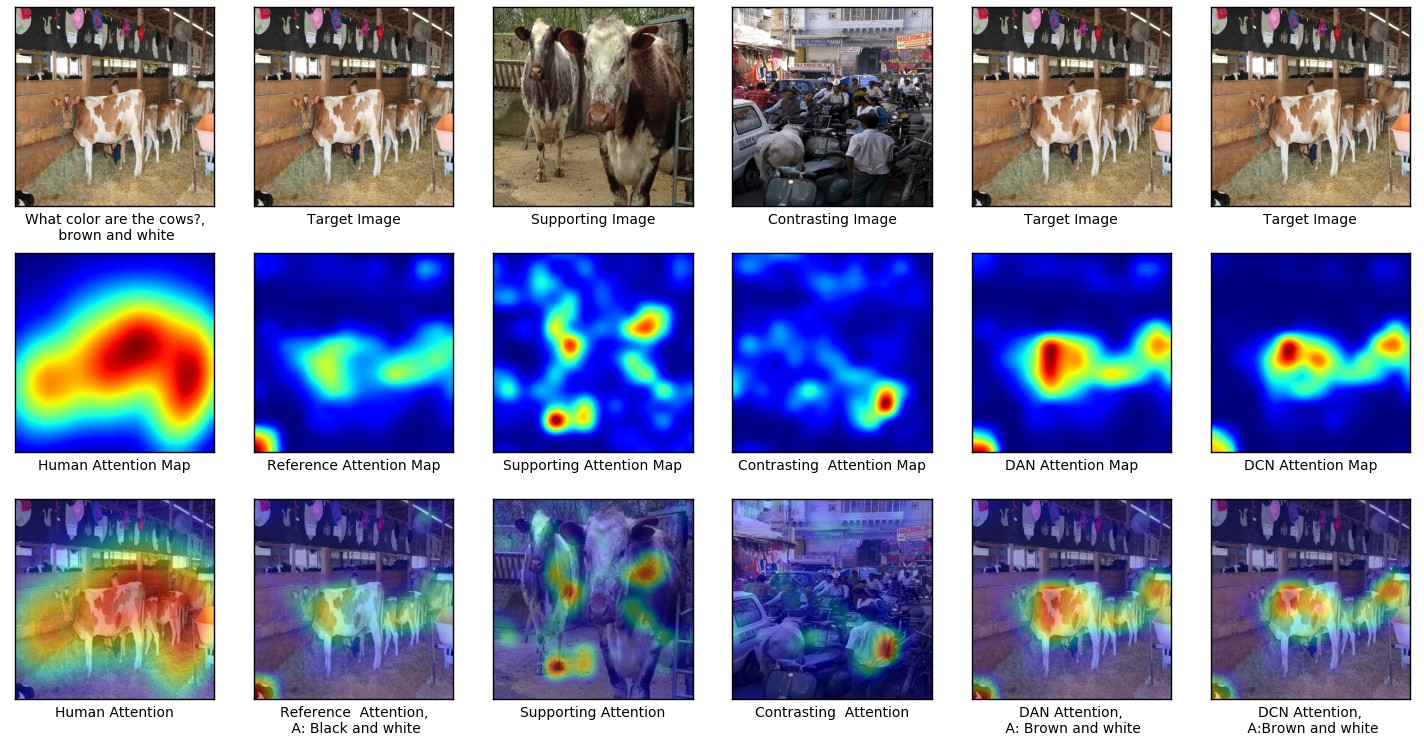}
	\caption{ In this figure, the first	row indicates the given target image, supporting image and opposing image. second row indicates the attention map for human\cite{Das_EMNLP2016}, reference attention map, supporting attention map , opposing attention map, DAN and DCN attention map respectively. Third row generates result by applying attention map on corresponding images.}
	\label{fig:attention}
\end{figure*}


	\subsection{Attention Visualization}
	The main aim of the proposed method is to obtain an improved attention that correlates better with human attention. Hence we visualize the attention regions and compare them. In attention visualization we overlay the attention probability distribution matrix, that is the most prominent part of a given image based on the query question. The procedure followed is same as that followed by Das {\it et al.} \cite{Das_EMNLP2016}. We provide the results of the attention visualization in figure~\ref{fig:att}. We obtain significant improvement in attention by using DCN as compared to the SAN method~\cite{Yang_CVPR2016}. Figure~\ref{fig:attention} provides how the supporting and opposing attention map helps to improve the reference attention using DAN and DCN.  We have provided more results for attention map visualization on the project website \footnote{project website: \url{https://badripatro.github.io/DVQA/} }.
	

	


\section{Discussion}
In this section we further discuss different aspects of our method that are useful for understanding the method in more detail

We first consider how exemplars improve attention. In differential attention network, we use the exemplars and train them using a triplet network. It is known that using a triplet (\cite{Hoffer_Springer2015} and earlier by \cite{Frome_ICCV2007}), that we can learn a representation that accentuates how the image is closer to the supporting exemplar as against the opposing exemplar. The attention is obtained between the image and language representations. Therefore the improved image representation helps in obtaining an improved attention vector. In DCN the same approach is used with the change that the differential exemplar feature is also included in the image representation using projections. More analysis in terms of understanding how the methods qualitatively improves attention is included in the project website. 

We next consider whether improved attention implies improved performance. In our empirical analysis we observed that we obtain improved attention {\it and} improved accuracies in VQA task. While there could be other ways of improving performance on VQA (as suggested by MCB\cite{Fukui_arXiv2016} ) these can be additionally incorporated with the proposed method and these do yield improved performance in VQA

Lastly we consider whether image (I) and question embedding (Q) are both relevant. We had considered this issue and had conducted experiments by considering I only, by considering Q only, and by considering nearest neighbor using the semantic feature of both Q\&I. We had observed that the Q\&I embedding from the baseline VQA model performed better than other two. Therefore we believe that both contribute to the embedding. 
\section{Conclusion}
In this paper we propose two different variants for obtaining differential attention for solving the problem of visual question answering. These are differential attention network (DAN) and differential context network (DCN). 
Both the variants provide significant improvement over the baselines. 
The method provides an initial view of improving VQA using an exemplar based approach. In future, we would like to further explore this model  and extend it to joint image and question based attention models.
\section{Acknowledgment}
We acknowledge the help provided by our Delta Lab members and our family
who have supported us in our research activity.
{\small
\bibliographystyle{ieee}
\bibliography{mybibfile.bib}
}
\appendix

\appendix

\section{Experimental Setup}
\label{sec:exp}

\subsection{Dataset}
	We have conducted our experiments on two types of dataset, first one is VQA dataset ,which contains human annotated question and answer based on images on MS-COCO dataset. Second one is HAT dataset based on attention map.  
	\subsubsection{VQA dataset}
	VQA dataset\cite{VQA} is one of the largest dataset  for VQA benchmark so far.  It built on complex images from ms-coco dataset. VQA dataset contains total 204721 images, out of which, 82783 images for training, 40504 images for validation and 81434 images for testing. Each image in the MS-COCO dataset\cite{Lin_ECCV2014} is associated with 3 questions and each question has 10 possible answers. This dataset is annotated by different people. So there are  248349 QA pair for training, 121512 QA pairs for validating and 244302 QA pairs for testing. 
	We use the top 1000 most frequently output as our possible answer set as is commonly used. This covers 82.67\%  of the train+val answer.

	\subsubsection{VQA-HAT(Human Attention) dataset}
	We used VQA-HAT dataset\cite{Das_EMNLP2016}, which is developed based on the de-blurring task to answering visual questions. 
	This dataset contains human attention map for training set of 58475 example out of 248349 VQA training set. It contains 1374 validation  example out of 121512 examples of question image pair in VQA validation set.

	\begin{algorithm*}
		
		\caption{Rank Correlation Procedure}\label{Correlation}
		\begin{algorithmic}[1]
			\Procedure{:}{Initialization}
			\State $P_{HAM}$: Probability distribution of Human Attention Map
			\State $P_{DAN}$	: Probability distribution of Differential Attention 
			\BState \emph{\textbf{Rank}}:
			\State Compute Rank of $P_{HAM}$ : $R_{HAM}$
			\State Compute Rank of $P_{DAN}$ : $R_{DAN}$ 
			
			\BState \emph{\textbf{Rank Difference }}:
			
			\State Compute difference in rank between $R_{HAM}$ \& $R_{DAN}$ : $Rank_{Diff}$
			\State Compute square of rank difference $Rank_{Diff}$ :$S_{Rank\_Diff}$

			\BState \emph{\textbf{Rank Correlation}}:
			\State Compute Dimension of $P_{DAN}$ :N
			\State Compute Rank Correlation using :
			\[R_{Cor}= 1- {\frac{6*S_{Rank\_Diff}}{N^3 - N}}\]
			
			\EndProcedure
		\end{algorithmic}
	\end{algorithm*}
	
	  \begin{figure*}[ht]
	\includegraphics[width=1.0\textwidth]{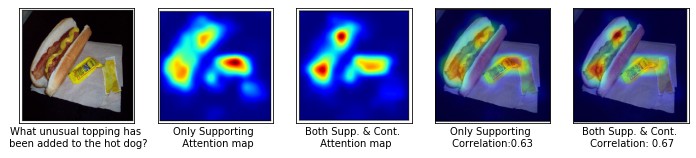}
	\caption{Importance of Supporting exemplar vs both. the first column in the figure indicates about image and corresponding question, the second  and third term indicates attention map for supporting exemplar and both supporting and opposing exemplar. The fourth and fifth column gives the value of rank correlation for supporting and both.}
	\label{fig:a1}
\end{figure*}
\subsection{Evaluation methods}
	\subsubsection{VQA dataset}
	VQA dataset contain 3 type of answer: yes/no, number  and other. 
	The evaluation is carried out using two test splits,i.e test-dev and test-standard. The question in corresponding test split are answered using two ways: Open-Ended\cite{VQA} and Multiple-choice. Open-Ended task should generate a natural language answer in form of single word or phrase. For each question there are 10 candidate answer provided with their respective confidence level. Our module generates a single word answer on the open ended task. This answer can be evaluated using accuracy metric provide by Antol {\it et al.}\cite{VQA} as follows.

	\begin{equation} 
	Acc=\frac{1}{N}\sum_{i=1}^{N}{min\big(\frac{\sum_{t \in T^{i} }{\mathbb {I}[a_i=t]}}{3},1 \big)}
	\end{equation}
	
	Where $a_i$ the predicted answer and t is the annotated answer in the target answer set $T^{i}$  of the $i^{th}$ example and $\mathbb {I}[.]$ is the indicator function.
	The predicted answer $a_i$ is correct if at least 3 annotators agree on the predicted answer. If the predicted answer is not correct then the accuracy score depends on the number of annotator that agree on the answer. Before checking accuracy, we need to convert the predicted answer to lowercase, number to digits and punctuation \& article to be removed. 	 	

	\subsubsection{HAT dataset}
	We used rank correlation technique to evaluate\cite{Das_EMNLP2016} the correlation between human attention map and DAN attention probability. Here we scale down human attention map to 14x14 in order to make same size as DAN attention probability. We then compute rank correlation using the following steps. 
	Rank correlation technique is used to obtain the degree of association between the data.  The value of rank correlation\cite{Mcdonald_handbook2009} lies between +1 to -1. When $R_{Cor}$ is close to 1, it indicates positive correlation between them, When $R_{Cor}$ is close to -1, it indicates negative correlation between them, and when $R_{Cor}$ is close to 0, it indicates No correlation between them. A higher value of rank correlation is better.
	
	\subsection{ Training  and Model Configuration}
	We  trained the  differential attention model using joint loss in an end-to-end manner. We have used RMSPROP  optimizer to update the model parameter and configured hyper-parameter values to be as follows: {learning rate =0.0004 , batch size = 200, alpha = 0.99 and epsilon=1e-8} to train the classification network . In order to train a triplet model, we have used RMSPROP to  optimize the triplet model model parameter and configure hyper-parameter values to be: {learning rate =0.001 , batch size = 200, alpha = 0.9 and epsilon=1e-8}. We have used learning rate decay to to decrease the learning rate on every epoch by a factor given by:
	\[Decay\_factor=exp\left(\frac{log(0.1)}{a*b} \right)\] where value of  a=1500 and b=1250 is set empirically. Selection of training controlling factor($\nu$) has a major role during training. If $\nu$=1 , means updating  triplet and classification network parameter at a same rate . If $\nu$ $\gg$ 1 , means updating triplet net more frequently as compare to classification net. Since, triplet loss decreases much lower then classification loss, we fixed value of $\nu$ $\gg$ 1 that is a fixed value of $\nu$=10.

	\begin{table*}[ht]
		\caption{Rank Correlation on HAT Validation Dataset for DAN and DCN}
		\label{HAT_rank_correlation1}
		\begin{center}		
			\begin{tabular}{ lcccr } \hline
				
				\textbf{Models}  & \textbf{Val1} & \textbf{ Val2} & \textbf{Val3 }& \textbf{Val} \\ \hline 
				DAN (K=1)  & {0.3147}& { 0.2772}& { 0.2958}& { 0.2959} \\ 
				 
				DAN (K=2)  & {0.3280}& { 0.2933}& { 0.3057}& { 0.3090} \\ 
				DAN (K=3)  & {0.3297}& { 0.2947}& { 0.3058}& { 0.3100} \\ 
				DAN (K=4)  &\textbf {0.3418}&\textbf{ 0.3060}	&\textbf { 0.3133}& \textbf{ 0.3206} \\ \hline
						
				DCN Add\_v1(K=1)      & { 0.3172}& { 0.2783}& { 0.2968}	& { 0.2974} \\ 
				DCN Add\_v2(K=1)  & { 0.3186}& { 0.2812}& { 0.2993}	& { 0.2997} \\ 
				DCN Mul\_v1(K=1)      & {0.3205}&  { 0.2847}& { 0.3023}&  { 0.3025} \\
				DCN Mul\_v2(K=1)  & \textbf{ 0.3227}& \textbf{ 0.2871}& \textbf{ 0.3059}	& \textbf{ 0.3052} \\ \hline

				DCN Add\_v1(K=4)      & {0.3426}& {0.3058}&{ 0.3123}& { 0.3202} \\ 			 	
				DCN Add\_v2(K=4)  & {0.3459}&{ 0.3047}&{ 0.3140}& { 0.3215} \\ 
				DCN Mul\_v1(K=4)      & {0.3466}& {0.3059}&{ 0.3163}& { 0.3229} \\
				DCN Mul\_v2(K=4)  & \textbf{0.3472}&\textbf{ 0.3068}	& \textbf{ 0.3187}& \textbf{ 0.3242}  \\ \hline
				DAN (K=1,Random )  & {0.1238}&{ 0.1070}	& { 0.1163}& { 0.1157} \\ 
				DAN (K=5)  & {0.2634}&{ 0.2412}	& { 0.2589}& { 0.2545} \\
			\hline
					
			\end{tabular}	
		\end{center}
	\end{table*}
	\begin{figure*}[ht]
	\includegraphics[width=1.0\textwidth]{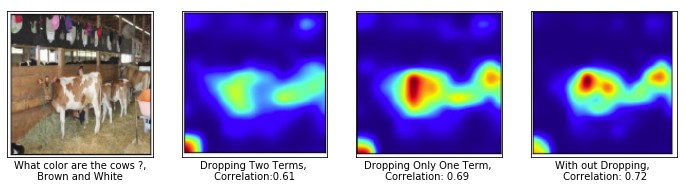}
	\caption{ Ablation Results for Dropping  terms in equation 3 and 4. The first column indicate the target image and its question, The second column provides the attention map \& rank correlation by dropping $2^{nd}$ in equation 3 \& $i^{st}$  term in equation 4. The third column gives the attention map \& rank correlation by dropping only  $i^{st}$  term in equation 4. Final column provides the attention map \& rank correlation by consider every thing in both the equation.}
	\label{fig:a2}
\end{figure*}

\begin{figure*}[ht]

	\centering
	\includegraphics[width=1.05\textwidth]{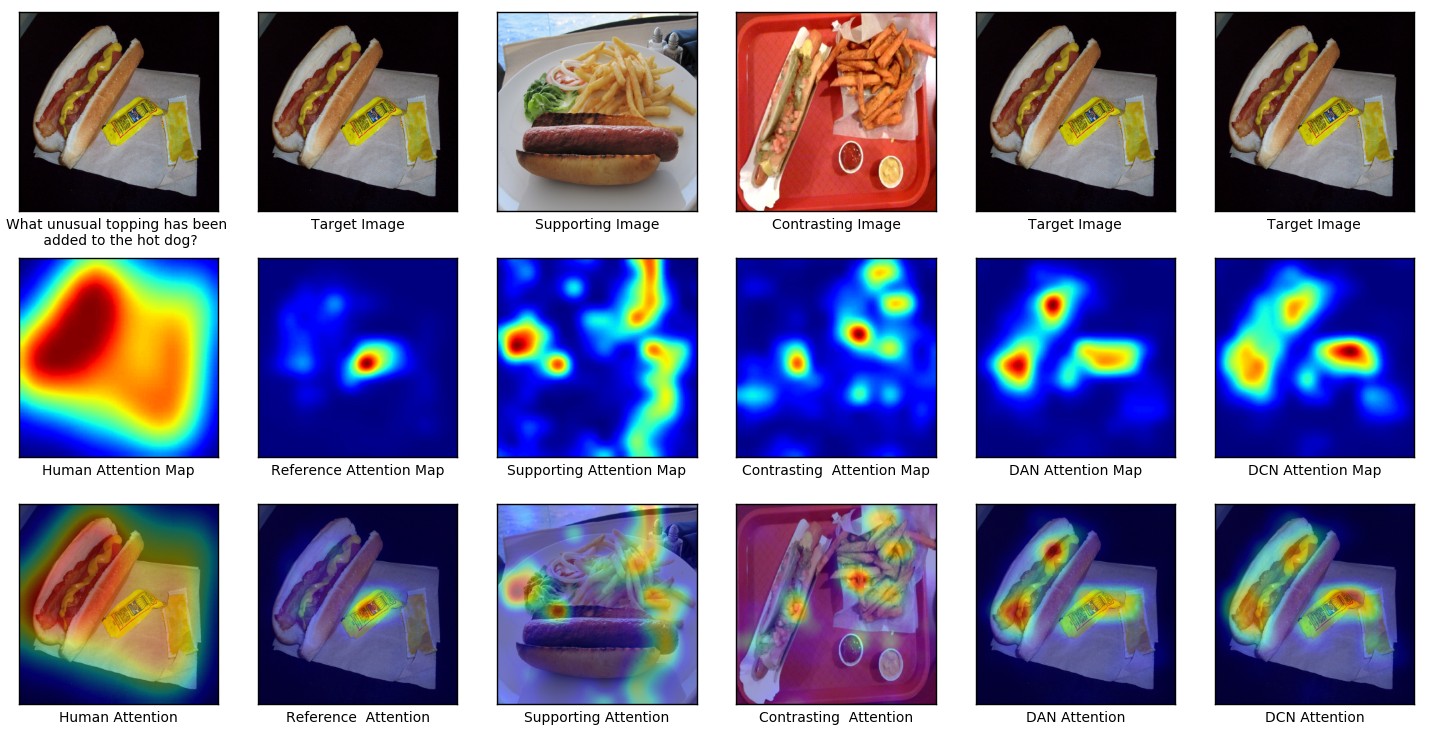}
	\caption{In this figure, the first	raw indicates given target image, supporting image and opposing image. second raw indicate the attention map for human\cite{Das_EMNLP2016}, reference attention map, supporting attention map , opposing attention map, DAN and DCN attention map respectively. Third raw generate result by applying attention map on corresponding images.}
	\label{fig:res6}
\end{figure*}

\begin{figure*}[ht]
	\centering
	\includegraphics[width=1.05\textwidth]{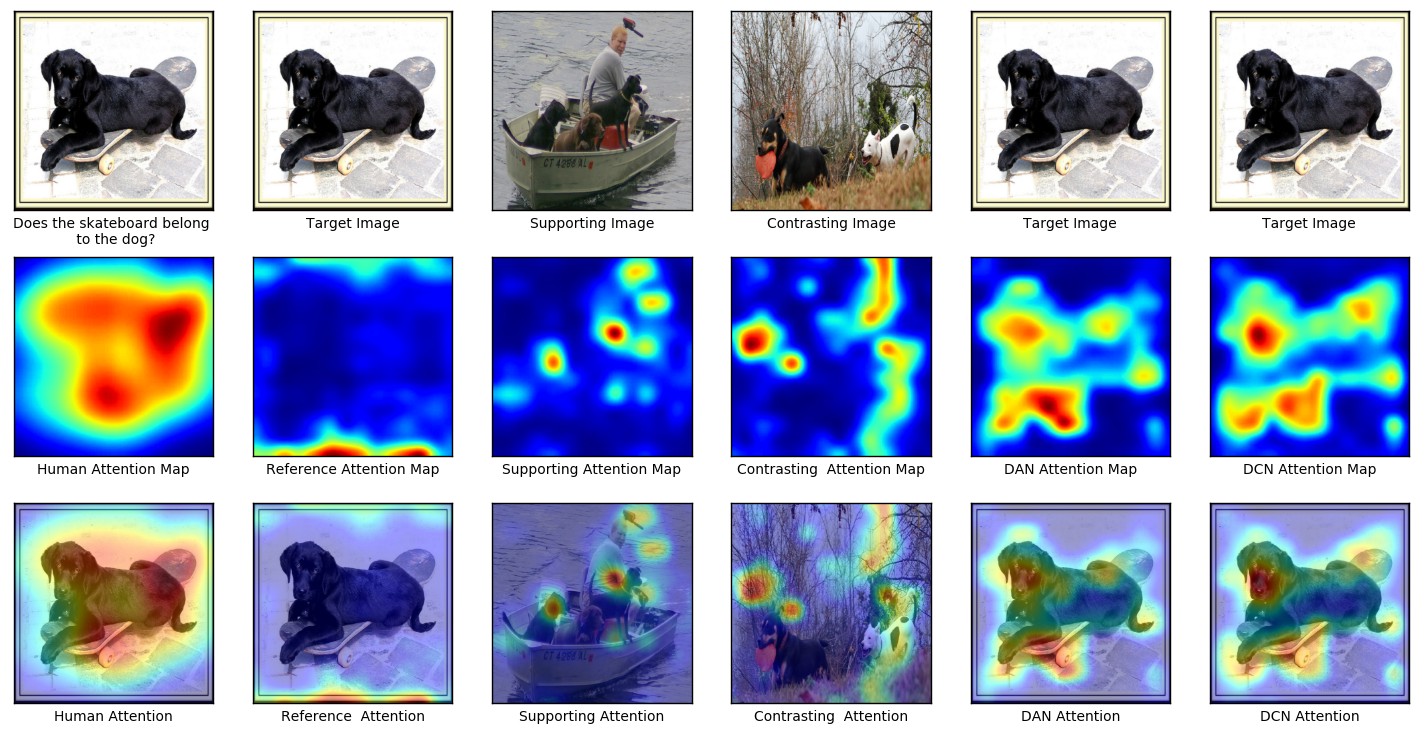}
	\caption{In this figure, the first	raw indicates given target image, supporting image and opposing image. second raw indicate the attention map for human\cite{Das_EMNLP2016}, reference attention map, supporting attention map , opposing attention map, DAN and DCN attention map respectively. Third raw generate result by applying attention map on corresponding images.}
	\label{fig:res7}
	
\end{figure*}
 \section{Results}
 In this section we provide additional results that were omitted from the main paper due to space limitation. 
\subsection{Rank Correlation results for DAN and DCN}
	In this sub-section we provide a few additional columns that were omitted from the results table in the main paper. Table-~\ref{HAT_rank_correlation1} provides the statistics of rank correlation on HAT validation dataset for various  differential attention networks(DAN) and differential context network(DCN). DAN network is varied by varying the number of nearest supporting and opposing exemplars. We did experiments by considering K=1,2, 3,4,5 and random selections of nearest and farthest neighbors.
	
	This table also mentions rank correlation for two types of DCN, i.e DCN Add and DCN Mul. Each type of network has two different methods for training , one is fixed scaling weights,i.e DCN Mul and second one is learn-able scaling weights,i.e, DCN Mul\_v1. From the statistics of rank correlation in this table indicates that learnable scaling weights performs better than fixed weights. Further, we observed that Multiplication network performs better than addition network in case of differential context. We did experiments for K=1,2,3,4, but this table only shows the results of K=1 and K=4 for number of nearest and farthest neighbors selections.

	We have computed rank correlation between attention probability of differential netwrok(DAN or DCN) and Human attention\cite{Das_EMNLP2016} provided by HAT for validation set.  This table contains 3 rank correlations for 3 attention maps per image on HAT validation dataset. First attention map gives better accuracy than other two. Finally we take an average of three rank correlation for a particular model. We can observe that, all our model attention maps correlate positively with human attention.

\subsection{How important are the supporting and contrasting exemplar?}
 We carried out an experiment by considering only the supportive exemplar in triplet loss mentioned in equation-2 and obtained consistent result as shown in figure~\ref{fig:a1}. From the rank correlation result, we can conclude that, If we use only the supportive exemplar, we obtain most of the gain in the performance. The quantitative results for this ablation analysis is shown in the table ~\ref{DAN_rank_correlation}, which provides  the rank correlation on HAT Validation Dataset.

  \begin{table}[ht]
	\caption{Rank Correlation for only Supporting Exemplar}
	\label{DAN_rank_correlation}
	\begin{center}		
		\begin{tabular}{ lc } \hline
			\textbf{Models} & \textbf{Rank-correlation} \\ \hline 
			DAN (K=4) +LQIA	 & { 0.312 $\pm$ 0.001} \\
            DAN (K=4) +MCB & { 0.320 $\pm$ 0.001} \\ \hline 
		\end{tabular}	
	\end{center}
\end{table}

 \subsection{Contribution of different term in DCN}
 We carried out an experiment by dropping the vector projection of $s_i^-$ on $s_i$ term in the supporting context $r_i^+$ as mentioned in equation-3 and the vector rejection of $s_i^+$ on $s_i$ term in opposing context $r_i^-$ as mentioned in equation-4 and obtained consistent result as shown in  figure~\ref{fig:a2}.
 The contribution of these terms in the corresponding equations are very small.The quantitative results for this ablation analysis is shown in the table ~\ref{DCN_rank_correlation}, which provides  the rank correlation on HAT Validation Dataset.

  \begin{table}[ht]
	\caption{Rank Correlation by Dropping various terms in DCN}
	\label{DCN_rank_correlation}
	\begin{center}		
		\begin{tabular}{ lc } \hline
			\textbf{Models} & \textbf{Rank-correlation} \\ \hline 
			DCN Mul\_v2(K=4) +LQIA	& \textbf{ 0.319$\pm$ 0.001} \\
            DCN Mul\_v2(K=4) +MCB & \textbf{ 0.3287$\pm$ 0.001}  \\ \hline
		\end{tabular}	
	\end{center}
\end{table}

\subsection{ Attention Visualization DAN and DCN with Supporting and Opposing  Exemplar }
The first row of figure-~\ref{fig:res6} indicates the target image along with a supporting and opposing image.  Second row provides human attention map, reference, supporting, opposing, DAN and DCN attention map respectively. Third row gives corresponding attention visualization for all the images. We can observe that from the given the target image and question: "what unusual topping has been added to the hot dog" , the reference model provides attention map($3^{rd}$ row, $2^{nd}$ column of figure-~\ref{fig:res6}) somewhere in the yellow part which is different from the ground truth human attention map ($3^{rd}$ row, $1^{st}$ column of figure-~\ref{fig:res6}). With the help of supporting and contrasting exemplar attention map($3^{rd}$ row, $3^{rd}$ \& $4^{th}$ column of figure-~\ref{fig:res6}), the reference model attention is improved, which is shown in DAN and DCN ($3^{rd}$ row, $5^{th}$ \& $6^{th}$ column of figure-~\ref{fig:res6}). The attention map of DCN model is more correlated with the ground truth human attention map than reference model. Thus we observe that with the help of supporting and contrasting exemplar, VQA accuracy is improving. Also, figure- ~\ref{fig:res7} provides attention visualization for DAN and DCN with the help of supporting and contrasting attention.




\subsection{Attention visualization of DCN with various Human Attention Maps }
\begin{figure*}[ht]

	\centering
	\includegraphics[width=0.8\textwidth]{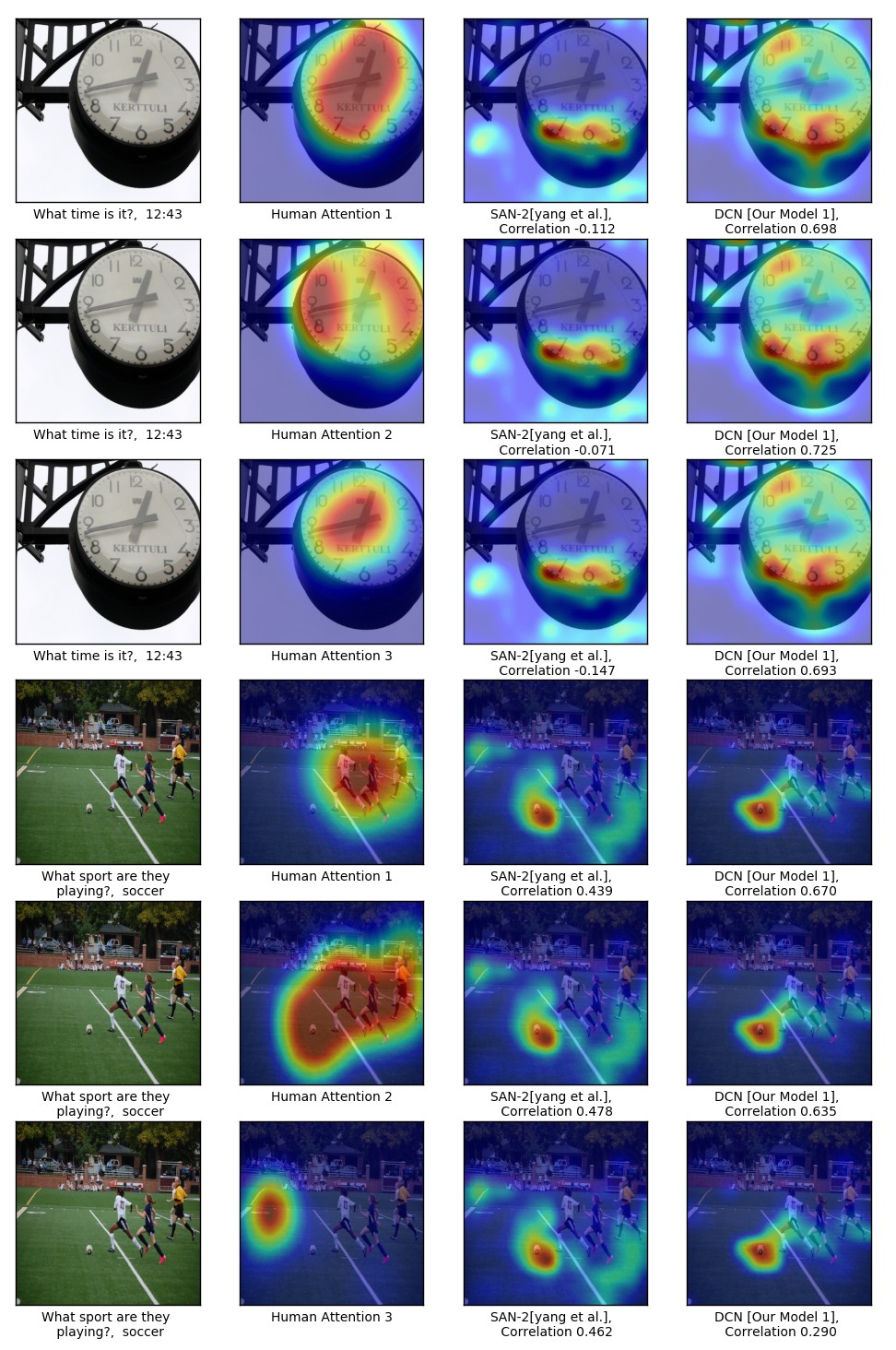}
	\caption{DCN Attention Map with  all ground truth three human attentions. The first row provides results for the human attention map-1 for HAT dataset\cite{Das_EMNLP2016}. The second row and third row  provides results for the human attention map-2 and human attention map-3. Similar results for another example. }
	\label{fig:res3}
\end{figure*}

\begin{figure*}[ht]
	\centering
	\includegraphics[width=0.8\textwidth]{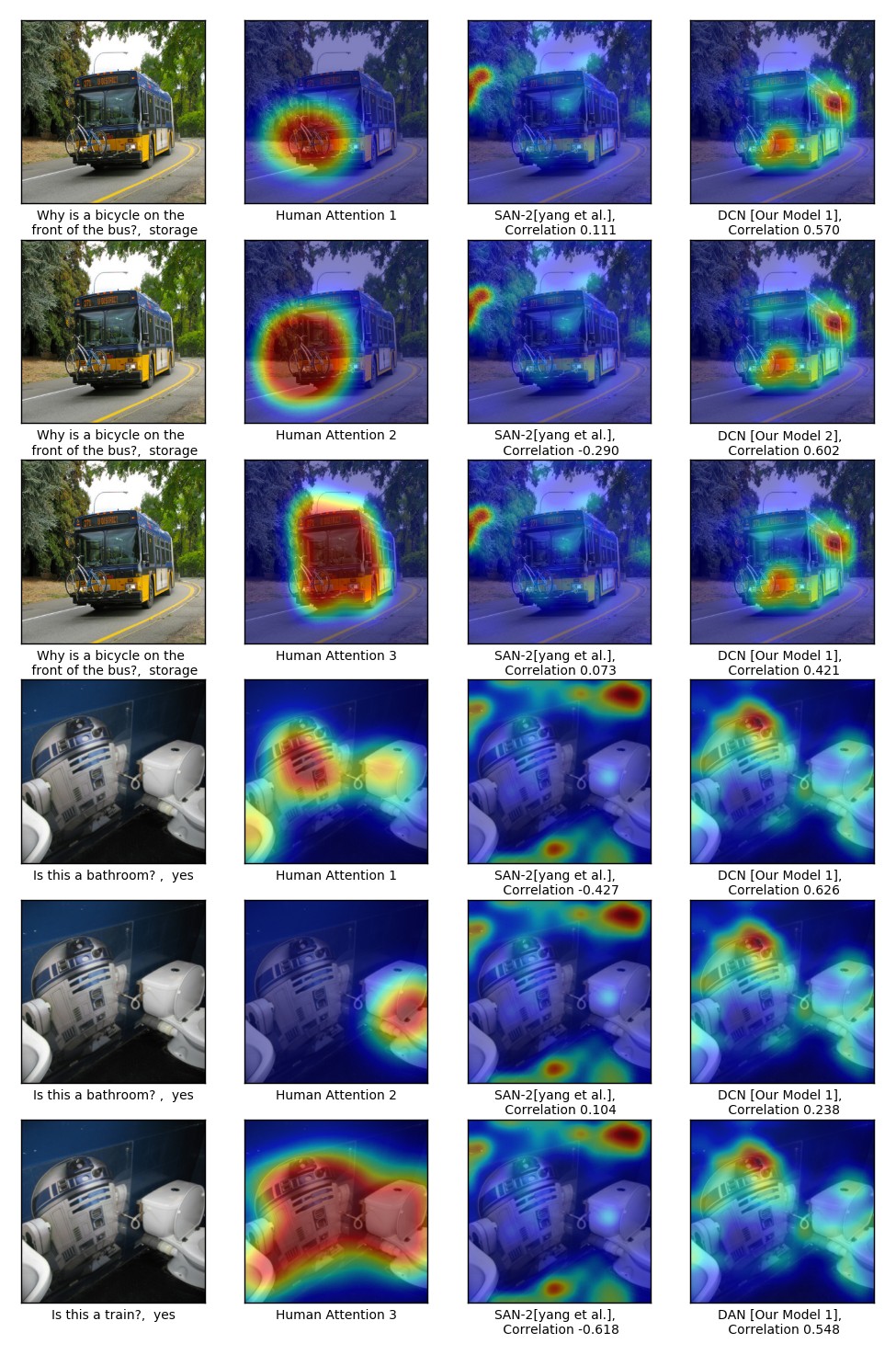}
	\caption{DCN Attention Map with  all ground truth three human attentions. The first row provides results for the human attention map-1 for HAT dataset\cite{Das_EMNLP2016}. The second row and third row  provides results for the human attention map-2 and human attention map-3. Similar results for another example.}
	\label{fig:res4}
\end{figure*}

We compute  rank correlation for all three ground truth human  attention map provide by VQA- HAT\cite{Das_EMNLP2016} val dataset with our DAN and DCN exemplar model and also visualized attention map with all thee ground truth human attention map as shown in figure~\ref{fig:res3} and ~\ref{fig:res4}. We can evaluated our rank correlation for all three human attention map and observed that human attention map one is better than attention map 2 and 3 in term of visualization and rank correlation as mention in figure~\ref{fig:res3} and ~\ref{fig:res4}. 

\subsection{Attention Visualization of DAN and DCN }
\begin{figure*}[ht]
	\centering
	\includegraphics[width=1.0\textwidth]{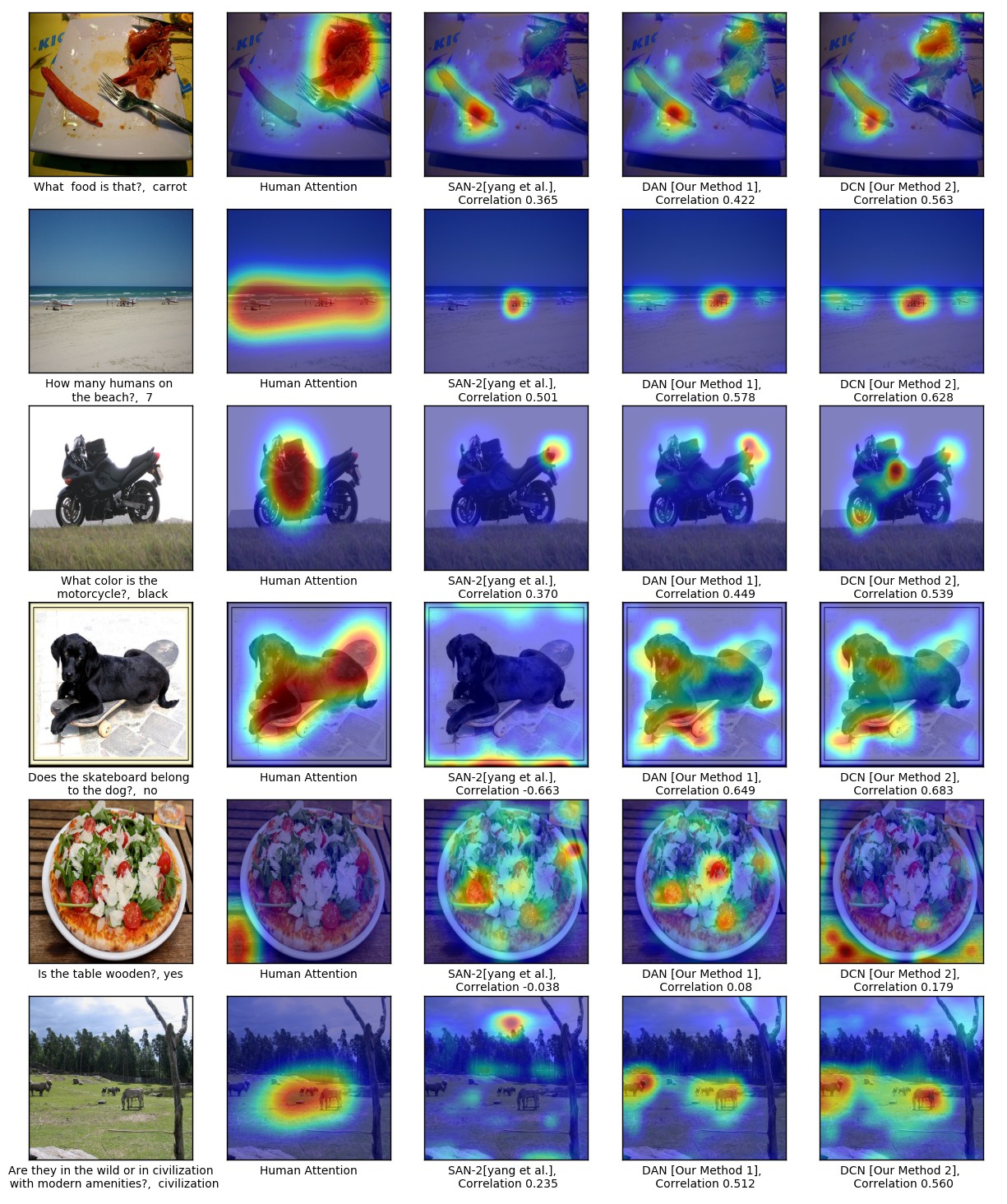}
	\caption{Attention Result for DAN and DCN. In this figure, the first column indicates target question and corresponding image, second column indicates reference human attention map in HAT dataset, third column refer to generated attention map for SAN, fourth column refers to rank correlation of our DAN model and final column refers to rank correlation for our DCN model.}
	\label{fig:res1}
\end{figure*}

\begin{figure*}[ht]
	\centering
	\includegraphics[width=0.95\textwidth]{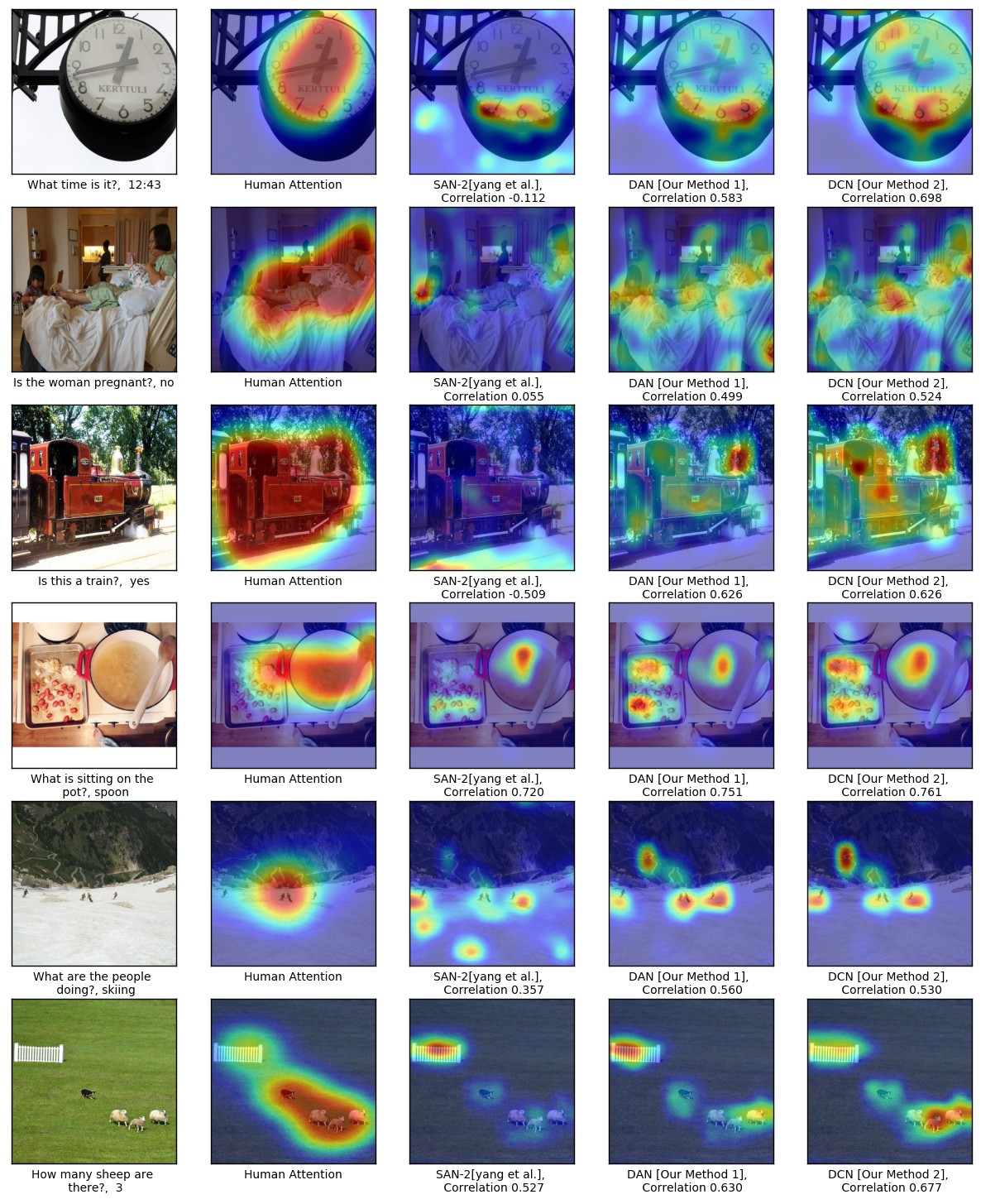}
	\caption{Attention Result for DAN and DCN,In this figure, the first column indicates target question and corresponding image, second column indicates reference human attention map in HAT dataset, third column refer to generated attention map for SAN, fourth column refers to rank correlation of our DAN model and final column refers to rank correlation for our DCN model.}
	\label{fig:res2}
\end{figure*}

We provide the results of the attention visualization in figure~\ref{fig:res1} and ~\ref{fig:res2}.  As can be observed in figure~\ref{fig:res1} and ~\ref{fig:res2}, we obtain significant improvement of rank correlation in attention map by using exemplar model(DCN or DAN) as compared to the SAN method~\cite{Yang_CVPR2016}. We can observed that DAN and DCN has more correlation with human attention.   We observed that DAN and DCN has better rank correlation then SAN attention map.

\section{Algorithm for Differential Attention}
The algorithm ~\ref{algo_attention}  for differential attention illustrates the dimensions of inputs and outputs.
\begin{algorithm*}[ht]
	\caption{Attention Mechanism}\label{algo_attention}
	\begin{algorithmic}[1]
		
		\Procedure{:Model}{$g_{i}$,$g_{s}$,$g_{c}$,$f_{i}$}
		\BState\emph{ Compute Attention Maps}:
		\State $s_i=ATTENTION\_MAP(g_{i},f_{i}) $, //dimension:1x196
		\State $s^+_i=ATTENTION\_MAP(g_{s},f_{i})$, //dimension:1x196
		\State $s^-_i=ATTENTION\_MAP(g_{c},f_{i})$, //dimension:1x196
		
		\BState\emph{If DAN Model}:
		\State $Loss\_Triplet= triplet\_loss(s_i,s^+_i,s^-_i)$
		\State $P_{att}= s_i$, //dimension:1x196
		
		\BState \emph{If DCN Model}:
		\State Compute Context: $r^+_i,r^-_i$ as in eq-3,4, //dimension:1x196
		\State $d_i=s_i \sym tanh(W_1 r^+_i- W_2 r^-_i )$, //dimension:1x196 
		\State $P_{att}= d_i$, //dimension:1x196
		\BState \emph{Compute Img \& Ques Attention  }:
		\State $V_{att}= \sum_{i}{P_{att}(i)}{G_{imgfeat}(i)}$, //dimension:1x512
		\State $A_{att}= V_{att} + f_{i}$, //dimension:1x512
		\State $Ans=softmax({W_A} A_{att} + b_A)$, //dimension:1x1000
		\EndProcedure
		\State {-----------------------------------------------------------------------------------------}		
		
		\Procedure{:Attention\_map}{$g_{i}$,$f_{i}$}
		\State $g_{i}$:Image feature, //dimension:14x14x512
		\State $f_{i}$: Question feature, //dimension:1x512
		
		\BState \emph{Match dimension}:
		\State $G_{imgfeat}$:Reshape  $g_{i}$ to 196x512 :$reshape(g_{i})$
		\State  $F_{quesfeat}$: Replicate $f_i$ to 196 times: $clone(f_{i})$ 
		\BState \emph{Compute Attention Distribution}:
		\State $h_{att}= tanh({W_I}{G_{imgfeat}} \Pexp ({W_Q}{F_{quesfeat}}+{b_q}))$
		\State $P_{vec}= softmax({W_P}{h_{att}}+{b_P})$, //dimension:1x196
		\State Return $P_{vec}$
		\EndProcedure
	\end{algorithmic}
\end{algorithm*}

\section{Details of Triplet and Quintuplet Network}
\subsection{Triplet Model}
	The concept triplet loss is motivated in the context of larger margin nearest neighbor classification\cite{Weinberger_JMLR2009}, which minimize the distance between target and supporting feature and maximize the distance between target and  contrasting feature. $f(x_{i})$ is the embedding feature of $i^{th}$ example of training image $x_{i}$ in n dimensional euclidean space.
	
	\begin{itemize}
		\item 	$f(s_{i})$ : The embedding of target
		\item	$f(s_{i}^{+})$ :The embedding of supporting exemplar
		\item	$f(s_{i}^{-})$ :The embedding of contrasting exemplar
	\end{itemize}
	
	The objective of triplet  loss is to make both supporting features   target will have same identity\cite{Schroff_CVPR2015} \& target and contrasting feature will have differ identity. which means  it brings all supporting features more close to target feature than that of contrasting features.
	\begin{equation}
	\begin{split}
	& D(f(s_{i}),f(s_{i}^{+})) +\alpha < D(f(s_{i}),f(s_{i}^{-}))\\
	& \forall{(f(s_{i}),f(s_{i}^{+}),f(s_{i}^{-}))} \in T
	\end{split}
	\end{equation}
	where $D(f(s_{i}),f(s_{j})) = ||f(s_{i})- f(s_{j})||_{2}^{2}$ is defined as the euclidean distance between $f(s_{i}) \& f(s_{j})$. $\alpha$ is the margin between supporting and contrasting feature.The default value of $\alpha$ is 0.2. T is training dataset set, which contain all set of possible triplets. 
	The objective function for  triplet loss is given by 
\begin{dmath}
T(s_i,s_i^+,s_i^-) = \texttt{max}(0, ||f(s_{i})-f(s_{i}^{+})||^{2}_{2} + \alpha - ||f(s_{i})-f(s_{i}^{-})||^{2}_{2})
\end{dmath}
	For simplicity ,the notation are replaced like this , $f(s_{i}) \rightarrow f , f(s_{i}^{+})\rightarrow f^+ , f(s_{i}^{-})\rightarrow f^-$.
	
	Gradient computation of L2 norm is given by  
	\begin{equation}
	\frac{\partial }{\partial x}{||f(x)||^{2}_{2}} = 2*f(x)\frac{\partial }{\partial x}{f(x)}
	\end{equation}

	The gradient of loss w.r.t the "Supporting" input $f^{+}$:
	\[
	\frac{\partial L}{\partial f^+} =\begin{cases}
	\Delta L^+, & \text{if $(\alpha +||f-f^{+}||^{2}_{2} - ||f-f^{-}||^{2}_{2})\ge 0$}\\
	0, & \text{otherwise}.
	\end{cases} 
	\]
	where $\Delta L^+=2*(f-f^+)\frac{\partial (f-f^+)}{\partial f^+}$

	\begin{equation}
	\frac{\partial L}{\partial f^+} =\begin{cases}
	-2(f-f^+), & \text{if $(\alpha +||f-f^{+}||^{2}_{2} - ||f-f^{-}||^{2}_{2})\ge 0$}\\
	0, & \text{otherwise}.
	\end{cases}  
	\end{equation}		
	
	The gradient of loss w.r.t the "Opposing" input $f^{-}$:
	\[
	\frac{\partial L}{\partial f^-} =\begin{cases}
	\Delta L^-, & \text{if $(\alpha +||f-f^{+}||^{2}_{2} - ||f-f^{-}||^{2}_{2})\ge 0$}\\
	0, & \text{otherwise}.
	\end{cases} 
	\]
	where $\Delta L^-=-2*(f-f^-)\frac{\partial (f-f^-)}{\partial f^-}$
	\begin{equation}
	\frac{\partial L}{\partial f^-} =\begin{cases}
	2(f-f^-), & \text{if $(\alpha +||f-f^{+}||^{2}_{2} - ||f-f^{-}||^{2}_{2})\ge 0$}\\
	0, & \text{otherwise}.
	\end{cases}  
	\end{equation}
	
	The gradient of loss w.r.t the "Target" input $f$:
\[
	\frac{\partial L}{\partial f} =\begin{cases}
	\Delta L, & \text{if $(\alpha +||f-f^{+}||^{2}_{2} - ||f-f^{-}||^{2}_{2})\ge 0$}\\
	0, & \text{otherwise}.
	\end{cases} 
\]
where $\Delta L=2*(f-f^+)\frac{\partial (f-f^-)}{\partial f} - 2*(f-f^-)\frac{\partial (f-f^-)}{\partial f}$

	\begin{equation}
	\frac{\partial L}{\partial f} =\begin{cases}
	2(f^--f^+), & \text{if $(\alpha +||f-f^{+}||^{2}_{2} - ||f-f^{-}||^{2}_{2})\ge 0$}\\
	0, & \text{otherwise}.
	\end{cases}  
	\end{equation}

	\subsection{Quintuplet Model}
	Unlike triplet model, In this model we considered two supporting and two opposing image along with target image. we have selected supporting and opposing image by clustering. i.e,	The 2000th nearest neighbor is divided into 20 cluster based on the distance from the target image. That is first cluster mean distance  is minimum cluster distance from target and 20th  cluster mean distance is the maximum cluster distance from the target. 
	\begin{itemize}
		\item 	$a_i=f(s_{i})$ : The embedding of Target 
		\item	$p_i^{+}=f(s_{i}^{+})$ :The embedding of supporting exemplar from cluster 1 
		\item	$n_i^{-}=f(s_{i}^{-})$ :The embedding of opposing exemplar from cluster 20 
		\item	$p_i^{++}=f(s_{i}^{++})$ :The embedding of supporting exemplar from cluster 2
		\item	$n_i^{--}=f(s_{i}^{--})$ :The embedding of opposing exemplar from cluster 19
	\end{itemize}
	
	The objective of quintuplet is to bring $p_{i}^{+}$(cluster 1) supporting feature more close to target feature than that of $p_{i}^{++}$ (cluster 2) supporting feature than that of $n_{i}^{--}$ (cluster 19) opposing feature than that of $n_{i}^{-}$(cluster 20) opposing feature. 
	
	\begin{equation}
		\begin{split}
	& D(a_{i},p_{i}^{+}) + \alpha_1 < D(a_{i},p_{i}^{++}) 
	+ \alpha_2 < \\
	& D(a_{i},n_{i}^{--})  + \alpha_3< D(a_{i},n_{i}^{-}), \\
    & \forall{(a_{i},p_{i}^{+},p_{i}^{++},n_{i}^{--},n_{i}^{-})} \in T
		\end{split}
	\end{equation}
	where $\alpha_1$,$\alpha_2$,$\alpha_3$ are the margin between $p_{i}^{+} \& p_{i}^{++} $ , $p_{i}^{++} \& n_{i}^{--} $, $n_{i}^{--} \& n_{i}^{-} $ respectively. T is training dataset set, which contain all set of possible quintuplet set. 
	
	Objective function for Quintuplet loss\cite{Huang_CVPR2016} is defined as :
	\begin{equation}
	\min{\sum^{N}_{i=1} (\varepsilon_{i} +\chi_{i}+\phi_{i})+\lambda||\theta||_{2}^{2}}\\
	\end{equation}
	subjected to : \\
	$  max(0,\alpha_1 + D(a_i,p_{i}^{+}) - D(a_i,p_{i}^{++})) \le \varepsilon_{i}\\	
	max(0,\alpha_2 + D(a_i,p_{i}^{++}) - D(a_i,n_{i}^{--})) \le \chi_{i}\\	
	max(0,\alpha_3 + D(a_i,n_{i}^{--})  - D(a_i,n_{i}^{-})) \le \phi_{i}$\\	
	$\forall{i},\varepsilon_{i} \ge 0, \chi_{i} \ge 0,\phi_{i} \ge 0$\\
	
	where $\varepsilon_{i},\phi_{i},\chi_{i}$ are the slack variable  and $\theta$  is the parameter  of attention network and $\lambda$ is a regularizing control  parameter.The value of  $\alpha_1$, $\alpha_2$, $\alpha_3$ are 0.006, 0.2,0.006 set experimentally.

\end{document}